\pdfoutput=1
\documentclass[journal]{IEEEtran}
\usepackage{amsmath,amsfonts}
\usepackage{algorithmic}
\usepackage{algorithm}
\usepackage{array}
\usepackage[caption=false,font=normalsize,labelfont=sf,textfont=sf]{subfig}
\usepackage{textcomp}
\usepackage{stfloats}
\usepackage{url}
\usepackage{verbatim}
\usepackage{graphicx}
\usepackage{subcaption}
\usepackage{caption}
\usepackage{cite}

\usepackage{tabularray}
\begin{document}
\begin{sloppypar}

\title{ReCon1M:A Large-scale Benchmark Dataset \\for Relation Comprehension in\\ Remote Sensing Imagery}
\author{
    Xian~Sun, ~\IEEEmembership{Senior Member,~IEEE,}
    \and Qiwei~Yan,
    \and Chubo~Deng, 
    \and Chenglong~Liu,
    \and Yi~Jiang,
    \and Zhongyan~Hou, 
    \and Wanxuan~Lu,
    \and Fanglong~Yao,
    \and Xiaoyu~Liu,
    \and Lingxiang~Hao,
    \and Hongfeng~Yu 
\thanks{
Corresponding author: Chubo Deng.

X. Sun, Q, Yan, C. Liu, Y. Jiang, and Z. Hou are with the Aerospace Information Research Institute, Chinese Academy of Sciences, Beijing 100190, China, the Key Laboratory of Network Information System Technology (NIST), Aerospace Information Research Institute, Chinese Academy of Sciences, Beijing 100190, China, the University of Chinese Academy of Sciences and the School of Electronic, Electrical and Communication Engineering, University of Chinese Academy of Sciences, Beijing 100190, China (e-mail: sunxian@aircas.ac.cn; yanqiwei22@mails.ucas.ac.cn; liuchenglong20@mails.ucas.ac.cn; jiangyi22@mails.ucas.ac.cn; houzhongyan23@mails.ucas.ac.cn).

C. Deng, W. Lu, F. Yao, X. Liu, L. Hao, and H. Yu are with the Aerospace Information Research Institute, Chinese Academy of Sciences, Beijing 100190, China, the Key Laboratory of Network Information System Technology (NIST), Aerospace Information Research Institute, Chinese Academy of Sciences, Beijing 100190, China (e-mail: dengcb@aircas.ac.cn; luwx@aircas.ac.cn; yaofanglong17@mails.ucas.ac.cn; liuxy@aircas.ac.cn; haolx@aircas.ac.cn; yuhf@aircas.ac.cn;).
}
}

% The paper headers
\markboth{Manuscript to IEEE TGRS}%
{Shell \MakeLowercase{\textit{et al.}}: Bare Demo of IEEEtran.cls for IEEE Journals}

\maketitle

\begin{abstract}
Scene Graph Generation (SGG) is a high-level visual understanding and reasoning task aimed at extracting entities (such as objects) and their interrelationships from images. Significant progress has been made in the study of SGG in natural images in recent years, but its exploration in the domain of remote sensing images remains very limited. The complex characteristics of remote sensing images necessitate higher time and manual interpretation costs for annotation compared to natural images. The lack of a large-scale public SGG benchmark is a major impediment to the advancement of SGG-related research in aerial imagery. In this paper, we introduce the first publicly available large-scale, million-level relation dataset in the field of remote sensing images which is named as ReCon1M. Specifically, our dataset is built upon Fair1M and comprises 21,392 images. It includes annotations for 859,751 object bounding boxes across 60 different categories, and 1,149,342 relation triplets across 64 categories based on these bounding boxes. We provide a detailed description of the dataset's characteristics and statistical information. We conducted two object detection tasks and three sub-tasks within SGG on this dataset, assessing the performance of mainstream methods on these tasks.
\end{abstract}


\section{Introduction}
\IEEEPARstart{T}{here} exists vast number of objects across the Earth, forming a complex interconnected system rather than isolated entities. These objects engage in diverse relations, such as explicit spatial relations and implicit semantic relations, etc. Accurately object relation comprehension, bridges the gap between objects detection and inference, is the critical process of cognition, which involves in knowing, learning and understanding objects. However, it is challenging to fully capture some relations by shooting at a horizontal angle in daily life because of wide-span relations (e.g. the imminent return of a ship and its corresponding port) and complex relations of multiple objects (e.g. the uninterrupted interplay among thousands of objects within a city). Remote sensing techniques offer a solution, enabling the acquisition of comprehensive, high-resolution surface data from a bird's-eye perspective. This facilitates the observation of diverse objects and provides solid data foundation for relation comprehension.

\begin{figure*}
\centering
\includegraphics[width=\linewidth]{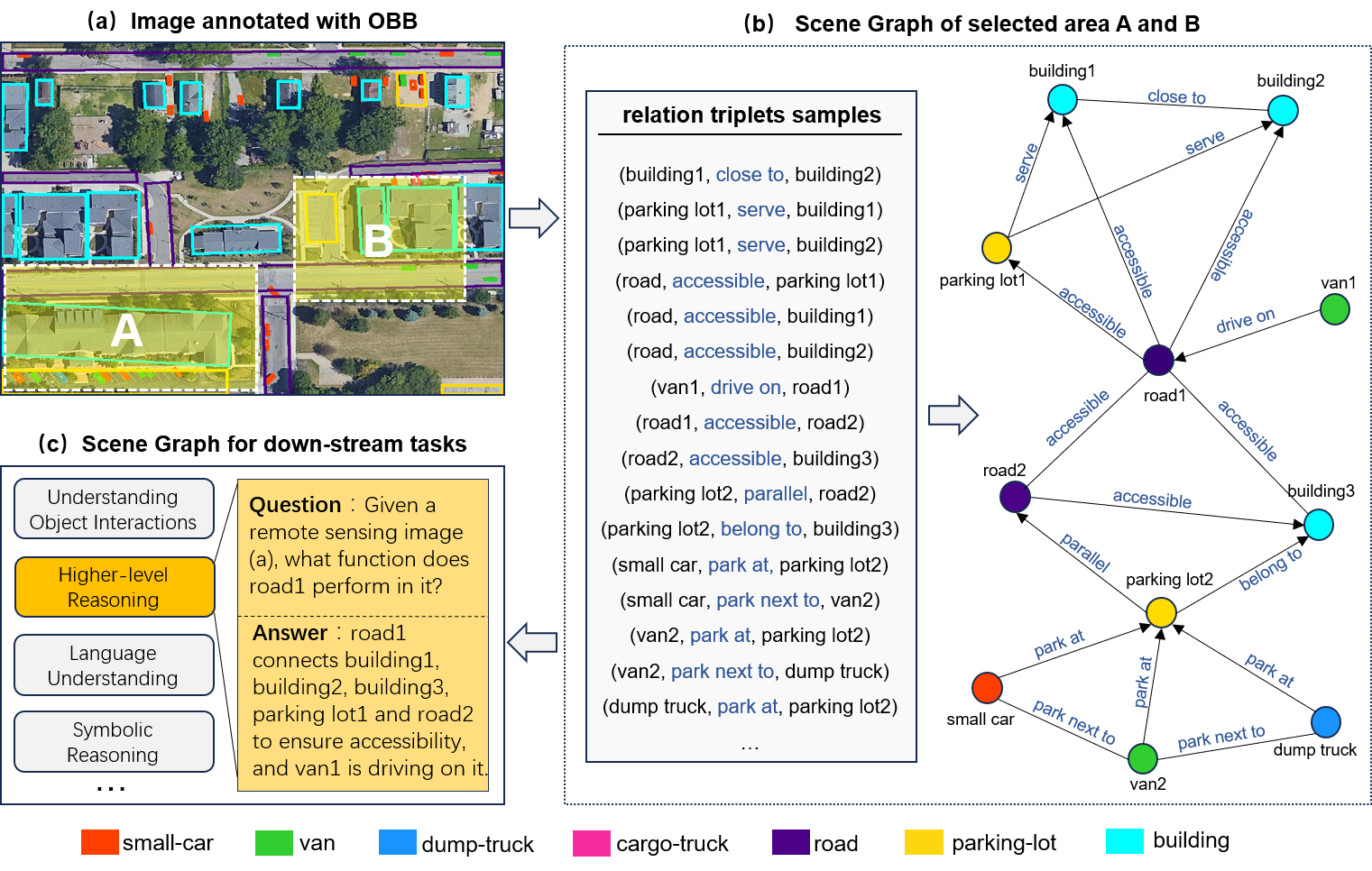}
\caption{Visualization example of the ReCon1M dataset. (a) shows the visualization of the oriented bounding box (OBB)annotations. Due to the complexity of the scene and the large number of objects and related instances, we selected regions A and B for scene graph visualization (as shown in (b)). Figure (c) shows several downstream tasks that scene graph can assist and gives a specific example of higher-order reasoning in visual question answering.}
\label{fig:intro_pic}
\end{figure*}

Currently, numerous researchers utilize remote sensing data for various object perception tasks, including classification \cite{xu2024attention,sun2022scan}, detection \cite{wang2019fmssd,sun2023revealing} and segmentation \cite{fu2018wsf,lu2021boundarymix}. The advent of artificial intelligence technology, particularly deep learning, has led to significant breakthroughs in this domain. Models based on remote sensing data can now proficiently recognize multiple objects within images, resulting in substantial improvements in accuracy \cite{sun2022ringmo}. Nevertheless, the majority of remote sensing models remain at the level of identifying objects, lacking the capability to understand the relations between objects, which is not enough to support deeper remote sensing cognition tasks. One primary contributing factor is the scarcity of large-scale, diverse remote sensing relation comprehension datasets.

A generous object relation dataset would greatly improve the generalization and accuracy of the object relation comprehension models. For instance, in the field of computer vision, the emergence of numerous object relation comprehension models has been propelled by renowned large-scale object relation datasets like Visual Genome (VG)\cite{r17}, Visual Relationship Detection (VRD)\cite{r16}. These datasets leverage natural images shooting at a horizontal angle to effectively comprehend the relations between objects. Similarly, to achieve a more profound level of cognitive understanding in remote sensing, there is a growing need for next-generation datasets to serve as training resources and benchmarks for object relation tasks specific to remote sensing applications.

As a result, in order to further advance research in remote sensing relation comprehension study, the following areas related to datasets need attention and enhancement.

\begin{itemize}
  \vspace{-1mm}
    \item 
    The size of the datasets needs to be scaled up in terms of the number of images and raltion annotations. Scene graph models often have a large number of parameters. To effectively tune these parameters and capture complex relations in data, a large dataset is needed. Without sufficient data, the model may not generalize well to unseen examples. Moreover, models learn hierarchical representations of features from raw data. To capture diverse and nuanced features, a large dataset is necessary to provide enough variation and examples. In general, more data leads to better performance, up to a certain point. As the dataset size increases, the model has more opportunities to learn and improve its performance on the task at hand.

    \item
    The number of relation category need to enrich. Complex scenes often involve a wide range of interactions and relation between objects. Having more relation types enables the model to adapt to these complexities and accurately represent the scene's content. More relation types allow for a detailed and nuanced representation of the interactions between objects in a scene. By incorporating diverse relation types, the model can generalize better to different scenes and scenarios. This helps improve the model's robustness and performance on a wide range of tasks and datasets, which benefit various tasks in computer vision and natural understanding.
    
    \item
    The importance of semantic relation should be emphasized. Spatial relations hold importance in tasks such as object localization, scene layout analysis, and certain visual reasoning tasks. While semantic relations provide a higher-level understanding of scenes, such as object categorization, scene understanding, and question answering. It captures the interactions and dependencies between objects based on their semantic meaning, such as "supply of", "part of," "belongs to," or "power." These relations provide crucial information about the functional and conceptual aspects of the scene, enabling a deeper understanding of the scene's content. 
    
    By incorporating semantic relations, models can leverage symbolic reasoning techniques to perform tasks that require logical inference and deduction. Semantic relations align well with symbolic reasoning, where logical relations are explicitly represented and manipulated. By understanding semantic relations, models can reason about the roles, functions, and attributes of objects in a scene. This can lead to sophisticated reasoning capabilities, such as understanding causality, temporal relations, and hierarchical structures within the scene. Therefore, incorporating a semantic relation in a scene graph can lead to more comprehensive, nuanced, and contextually rich representations of scenes, which can benefit various downstream tasks such as image scene understanding and visual reasoning.
\end{itemize}

\begin{figure*}[!t]
\centering
\includegraphics[width=0.95\textwidth]{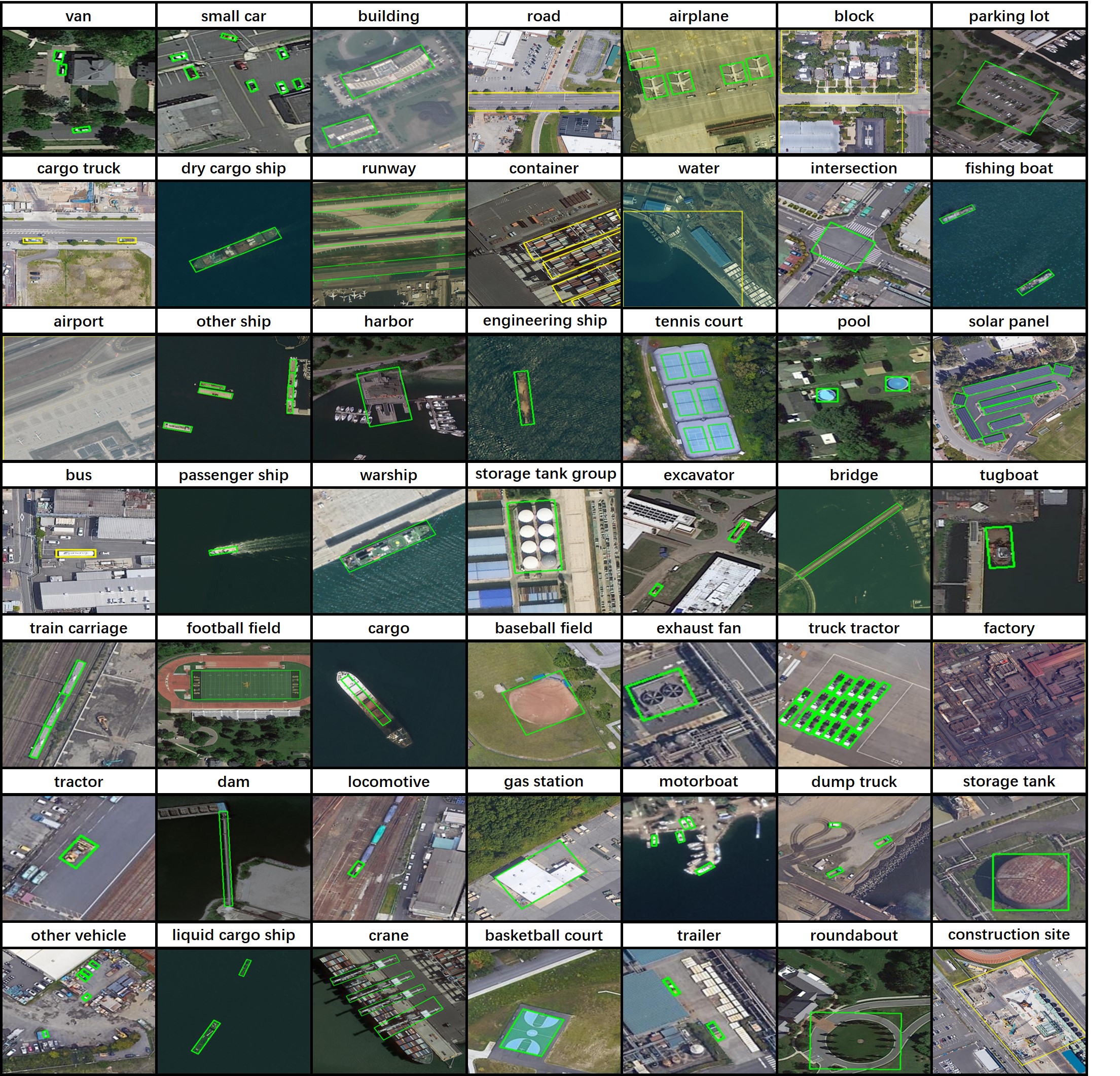}
\caption{Example images of some object categories in ReCon1M.}
\label{fig:obj_sample}
\end{figure*}

\begin{figure*}[!t]
\centering
\includegraphics[width=7in]{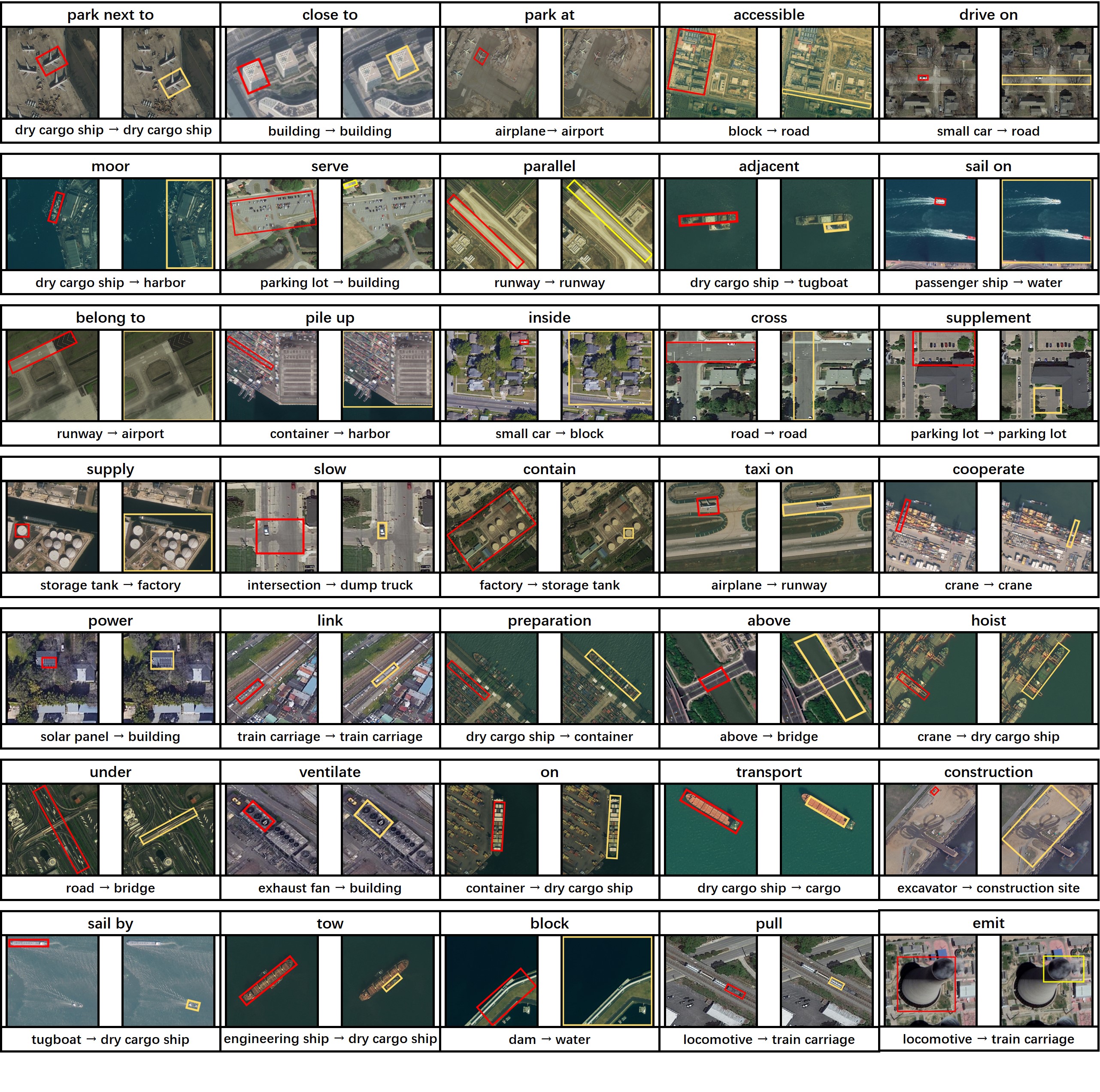}
\caption{Example instances of some relation categories in ReCon1M. A relation instance can be represented by a triplet $\langle subject, relation, object \rangle$. Each relation instance is represented by two images, where the left and right images respectively show the subject and the object, with the relation label above and the labels of the subject and object below the images}
\label{fig:rel_sample}
\end{figure*}
		
As a result, in order to resolve the issues mentioned above, we propose a novel benchmark dataset for object relation recognition in remote sensing field which is named as ReCon1M. Some representative examples are given in Fig.~\ref{fig:intro_pic}. In the ReCon1M dataset, the remote sensing imagery come from FAIR1M dataset, which is a high-resolution object recognition dataset. Base on the Fair1M, we label the relations between objects. Since many relations involve new objects that have not been annotated in the original dataset, we annotate these new objects with oriented bounding box under the guidance of many experts in remote sensing. To the best of our knowledge, ReCon1M is the largest object relation dataset suitable for remote sensing scenes. 

The annotation of object relation is more challenging task than object recognition annotation. Because it’s unavoidable that each annotator has recognition bias and sometime two different annotators give distinguish relation labels at the same image. More seriously, it’s unclear boundaries for annotation of some relations that makes it more challenge to accomplish. For example, it is hardly defined the boundary of relation “close to” quantitatively. Therefore, a unified annotation standard is established to overcome this problem and all annotator follow the same criterion. The standard is consistent during all annotation process. As the above example, we define if the distance between two objects is less than the longer side of the object and they are not adjacent to each other, than we annotate the relation between these two objects as “close to”.

In summary, the proposed ReCon1M benchmark dataset aims at providing a large-scale object relation dataset to the remote sensing community. With the support of ReCon1M, we hope a growing number of novel algorithms will emerge in the field of remote sensing image interpretation. The main contribution of this work is briefly summarized as follows:

\begin{itemize}
  \vspace{-1mm}
    \item 
    To the best of our knowledge, ReCon1M is the first benchmark dataset in the field of remote sensing imagery with a million-level relational annotation for Scene Graph Generation (SGG). The use of oriented bounding box annotations in ReCon1M is particularly suited to the characteristics of objects in remote sensing images. This dataset represents a significant advancement over existing datasets in terms of the number and categories of object annotations and the number and categories of relational annotations. 

    \item
    We have modified several representative SGG algorithms to accommodate SGG tasks with oriented bounding box annotations. We evaluated the performance of these algorithms on our dataset and analyzed the results. This evaluation provides a reference benchmark for the design of subsequent SGG algorithms in the field of remote sensing imagery. 

    \item
    Considering that ReCon1M is a dataset with OBB annotations used in remote sensing scenes, directly replacing the detectors in methods for natural scenes with rotated detectors does not yield satisfactory results. Therefore, we propose an efficient global context-Aware network to address the issues of OBB annotations, dense objects and relations, and large variations in object scales and distances in ReCon1M, aiming to improve the accuracy of relation prediction.
    
\end{itemize}

\begin{table*}[!t]
\centering
\caption{\\Statistics of the scene graph dataset in both common scene and remote sensing scene. "-" indicates that this attribute is not released}
\label{table:2D-image-datasets}
\begin{tblr}{
  colspec = {l l X[c] X[c] X[c] X[c] X[c] X[c]},
  rows = {m},  % 设置所有行的单元格内容垂直居中
  cell{1}{1,2,3} = {r=2}{m, c},  % 合并第一列和第二列的前两行，并垂直居中
  cell{3}{1} = {r=9}{m},
  cell{12}{1} = {r=3}{m},
  cell{1}{4} = {c=3}{c},
  cell{1}{7} = {c=2}{c},
  hline{1, 3, 12, Z} = {solid}, 
  vline{2, 3, 4, 7, 9} = {solid},
}
  scene  & Dataset  & Size & Objects &       &      & Relations & & Year \\
    &          &      & Annotation & \#bbox & \#categories & \#triplet & \#categories   &  \\
 common& Visual Phrase\cite{r36}      & 2,769     & HBB & 3,271     & 8      & 1,796     & 9      & 2011 \\
 & Scene Graph\cite{r19}        & 5,000     & HBB & 69,009    & 266    & 109,535   & 68     & 2015 \\
 & VRD\cite{r16}                & 5,000     & HBB & -         & 100    & 37,993    & 70     & 2016 \\
 & Visual Genome\cite{r17}      & 108,077   & HBB & 3,843,636 & 33,877 & 2,347,187 & 40,480 & 2017 \\
 & SpatialVOC2K\cite{r56}       & 2,026     & HBB & 5,775     & 20     & 9804      & 34     & 2018 \\
 & VrR-VG\cite{r53}             & 58,983    & HBB & 282,460   & 1,600  & 203,375   & 117    & 2019 \\
 & SpatialSense\cite{r55}       & 11,569    & HBB & -         & 3679   & 13,229    & 9      & 2019 \\
 & Open Images v4\cite{r52}     & 9,278,275 & HBB & 3,290,070 & 57     & 374,768   & 329    & 2020 \\
 & UnRel\cite{r54}              & 1,071     & HBB & -         & -      & 76        & 18     & 2020 \\
 remote sensing& RSSGD\cite{r15}              & -     & HBB & 13         & 39      & 76        & 18     & 2021 \\
 & S2SG\cite{r57}              & -     & MASK & -         & -      & 1200     & 12     & 2022 \\
 & ReCon1M(ours)                & 21,392    & OBB & 859,751   & 60     & 1,149,342 & 64     & 2024 \\
\end{tblr}
\end{table*}
		
\section{Related Work}
\subsection{Existing Scene Graph Dataset}
Datasets play an important role in the task of scene graph generation. Previously, datasets were predominantly focused on natural scenes. For instance, the VRD dataset \cite{r16} comprises approximately forty thousand relational patterns across one hundred object categories, encapsulated in five thousand images. Furthermore, the Visual Genome (VG) \cite{r17} has been extensively employed in applications such as visual question answering, captioning, and scene graph construction. It characterizes implicit relations by annotating connections between objects and their attributes, thereby transforming the context into feature-rich data vectors. Notably, Open Images \cite{r35} emerges as a substantial dataset, furnishing numerous instances for object detection and relational analysis. Additionally, the RW-SGD \cite{r19}, constructed on the basis of the Microsoft COCO dataset \cite{r4} and YFCC100m \cite{r37}, facilitates the comprehensive exploration of scene graphs and relational models by manually selecting 5000 images and leveraging Amazon Mechanical Turk (AMT) to generate specific scene configurations. Moreover, the HCVRD dataset \cite{r38}, an expansion of VRD, includes 1824 object classes, 52,855 images, 927 predicates, and 28,323 relations classes. For the generation of dynamic scene graphs, the CAD120 \cite{r16} and Action Genome (AG) \cite{r40} collectively provide a video action reasoning dataset that captures human social scenes and focuses on the temporal aspects of relation pattern analysis.

However, datasets based on relation patterns for remote sensing are currently scarce. The RSICD dataset \cite{r14} enables a detailed linguistic representation of remote sensing images. It stands out by providing extensive captioning data, which facilitates the examination of associated knowledge through the relations it generates. Additionally, Li et al. \cite{r15} introduced the Remote Sensing Image Scene Graph Dataset (RSSGD). They explored the construction of scene graphs for remote sensing images and incorporated global information using graph convolution networks, enhancing the analysis of spatial and relational data within these images.

\subsection{Existing Scene Graph Methods}
There is a lengthy research history behind the concept of building an in-depth understanding network based on relation patterns among objects. Semantic reasoning has been extensively studied in the realm of computer vision \cite{r1}, \cite{r2}, \cite{r3}, \cite{r4}, many innovative scholars have produced significant discoveries in the field of scene graph research based on semantic reasoning \cite{r6}, \cite{r7}. In the realm of Scene Graph Generation (SSG) methods, a distinction arises between statistically-informed methodologies and those rooted in deep learning.
Conditional Random Fields (CRF), a classical tool in this context \cite{r41}, exhibits proficiency in incorporating relational information into the models. In the SSG domain, CRF methods include two fundamental components: object detection and relation detection. Several refined SSG models, leveraging CRF principles, have showcased improved performance through the integration of robust relation prediction models and object detection models such as the Deep Relationship Network (DR-Net) \cite{r28} and the Semantic Compatibility Network (SCN) \cite{r47}. But there is a tendency to overlook the inherent order of relationships between two entities, leading to the confusion of subjects and objects. This limitation is particularly pronounced in tasks involving directed knowledge graph prediction, contributing to notable distortions in knowledge acquisition for relation prediction.
Knowledge graphs, representing and analysing information in scene graphs, manifest complex structures analogous to vector graph, wherein relations are distinctly defined through sets of entities and edges. Models grounded in TransE (Translation Embedding) have demonstrated exceptional utility in this context \cite{r8}. TransE conceptualizes relations as translations, treating the relation between a pair of objects as the translation of the attribute vector from one object to another, thereby acquiring optimal translation patterns.
 Drawing inspiration from TransE’s success in knowledge learning, recent investigations \cite{r8}, \cite{r48}, \cite{r10} explore the generation of visual connections through object features. Furthermore, methodologies like RLSV \cite{r49} and AT (Analogy Transfer) \cite{r50} simplify the knowledge vectors in scene graphs through transformations. However, this approach only focuses on the relation features between subjects and objects, neglecting the integration of contextual information. Consequently, numerous unresolved challenges persist within current TransE methodologies.
Beyond methodological challenges, there are significant defects in training data. Notably, the sparsity issue in predicate representations poses challenges, similar to the long-tail effect, complicating the task of visual relation detection. Using the Stanford VRD dataset \cite{r16} as an illustrative case, its potential predefined relations count exceeds 700k, highlighting that numerous real relations lack any training samples. UVTransE \cite{r51} introduces joint features for subject and object boundaries, effectively capturing contextual information. This class of analogy transfer-based SSG methods exhibits promising applications, and empirical findings substantiate their effectiveness.

\subsection{Limitation and Challenges in SGG}
\subsubsection{Long-Tail Problem in SGG}

Within the context of existing scene graphs, extant methodologies for relation recognition exhibit limitations in effectively addressing the prevalent long-tail problem. This issue causes from a non-uniform distribution of samples corresponding to identified relation patterns, leading to a pronounced concentration in the head and a flattened distribution in the tail. Consequently, most of the relation patterns in training data set suffer from an inadequacy of supporting data. The complexity of relations amplifies exponentially in scenarios involving multiple relations, and the reality of many visual relations in authentic scenarios being absent from training sets poses a challenge in achieving comprehensive cognition.

To overcome this challenge, current research in scene graph analysis endeavors to eliminate the consequences of the long-tail distribution. Noteworthy among these approaches are zero-shot and few-shot learning methodologies \cite{r20}, \cite{r21}, \cite{r22},\cite{r23}, \cite{r24}, strategically transferring relational knowledge to address long-tail positions, just like the principles of meta-learning. This strategic transfer enhances the model’s capacity to learn from limited samples.
Furthermore, the integration of prior knowledge methods introduces external information into the model. Specifically, introducing linguistic priors \cite{r25}, \cite{r26}, \cite{r16} and statistical priors \cite{r27}, \cite{r28}, \cite{r29}, \cite{r44} serves to improve the accuracy of recognizing relation. The alignment with prior knowledge contributes to an augmented scalability and heightened recognition capabilities of the model.

Moreover, from the standpoint of transfer learning, redistributing head relations to the tail emerges as a potential strategy for dealing the long-tail effect \cite{r31}. The Total Direct Effect (TDE) explores a training paradigm rooted in causal reasoning, where in a comparative analysis between the discovered causal relation network and the counterfactual network derived from the scene graph identifies and corrects errors in the transfer learning process \cite{r32}.

Despite previous datasets such as Open Images, Visual Phrase (VP), RW-SGD and HCVRD \cite{r35},\cite{r36},\cite{r19},\cite{r38} contain a wide range of relation classes, they also exhibit a long-tail distribution of infrequent relations, sharing the same issue of sparse data in their long tail region.

\subsubsection{Challenges in Remote Sensing Image Datasets}

Previous remote sensing datasets, including DIOR, HRRSD, and NWPU \cite{r11}, \cite{r12}, \cite{r13}, lack annotations for certain relation patterns, which makes it harder to model and compute the key relation in relation study. Therefore, creating a useful dataset based on relation patterns becomes essential when taking consideration of the features of remote sensing images and the complexity of relations. Relation-pattern-based remote sensing datasets are now very few. Because the spatial relations in RSICD dataset \cite{r14} are not well-represented in image captions, it is quite difficult to directly use this dataset for modeling and calculation. Semantic relation model and dataset for remote sensing scene understanding does not deeply analyze the characteristics of remote sensing images, and the feature relations in the dataset itself are insufficient, unable to cope with large-scale effective relation studies.

\begin{table}[t!]
\centering
\caption{\\The number of instances for each object category and dataset split.}
\label{tab:obj_category}
\begin{tblr}{
  colspec = {c X[c] X[c] X[c]},
  rows = {m},  % 设置所有行的单元格内容垂直居中
  row{1} = {font=\bfseries},
  hline{1-2, Z} = {solid}, 
  row{3-Z} = {rowsep=0.5pt}
}
Category & Train & Validation & Test \\
van                & 89820 & 29832 & 55716 \\
small car          & 83451 & 26879 & 54926 \\
building           & 68734 & 22700 & 45136 \\
road               & 35912 & 11859 & 23431 \\
airplane           & 17750 & 5917 & 12214 \\
block              & 15520 & 4862 & 9685 \\
parking lot        & 14749 & 4759 & 9399 \\
motorboat          & 14671 & 5259 & 8063 \\
dump truck         & 13210 & 4451 & 8391 \\
cargo truck        & 8755 & 3141 & 5398 \\
dry cargo ship     & 7459 & 2545 & 5841 \\
runway             & 7064 & 2404 & 4775 \\
container          & 6779 & 2372 & 4294 \\
water              & 4935 & 1606 & 3254 \\
intersection       & 4873 & 1555 & 3052 \\
fishing boat       & 4389 & 1310 & 1868 \\
other vehicle      & 3631 & 1285 & 2377 \\
storage tank       & 3494 & 1276 & 1969 \\
airport            & 3136 & 999 & 2143 \\
other ship         & 2583 & 875 & 1794 \\
harbor             & 2536 & 880 & 1636 \\
engineering ship   & 1518 & 639 & 1290 \\
tennis court       & 1817 & 527 & 1088 \\
pool               & 1666 & 573 & 1147 \\
solar panel        & 1850 & 500 & 926 \\
liquid cargo ship  & 1308 & 515 & 967 \\
crane              & 1288 & 474 & 721 \\
bus                & 949 & 255 & 916 \\
passenger ship     & 1023 & 354 & 719 \\
warship            & 1077 & 322 & 661 \\
storage tank group & 874 & 241 & 543 \\
excavator          & 773 & 173 & 709 \\
bridge             & 876 & 249 & 529 \\
tugboat            & 683 & 255 & 482 \\
basketball court   & 765 & 208 & 447 \\
trailer            & 691 & 253 & 448 \\
train carriage     & 679 & 228 & 405 \\
football field     & 620 & 191 & 366 \\
cargo              & 548 & 222 & 399 \\
baseball field     & 565 & 210 & 361 \\
exhaust fan        & 422 & 158 & 272 \\
truck tractor      & 305 & 129 & 311 \\
factory            & 322 & 104 & 202 \\
roundabout         & 239 & 92 & 166 \\
construction site  & 208 & 69 & 134 \\
chimney            & 176 & 57 & 72 \\
stadium            & 171 & 48 & 68 \\
smoke              & 154 & 44 & 59 \\
railway            & 124 & 31 & 47 \\
boarding bridge    & 96 & 39 & 42 \\
farmland           & 77 & 32 & 36 \\
helipad            & 68 & 21 & 31 \\
tractor            & 74 & 34 & 40 \\
greenbelt          & 32 & 14 & 18 \\
control tower      & 27 & 7  & 15 \\
dam                & 30 & 12 & 17 \\
typhoon spiral     & 25 & 5 & 13 \\
typhoon eye        & 23 & 3 & 9 \\
locomotive         & 22 & 18 & 13 \\
gas-station        & 6 & 1 & 5 \\
\end{tblr}
\end{table}

Drawing inspiration from TransE's success in knowledge learning, recent investigations \cite{r8}, \cite{r48}, \cite{r10} explore the generation of visual connections through object features. Furthermore, methodologies like RLSV \cite{r49} and AT (Analogy Transfer) \cite{r50} simplify the knowledge vectors in scene graphs through transformations. However, this approach only focuses on the relation features between subjects and objects, neglecting the integration of contextual information. Consequently, numerous unresolved challenges persist within current TransE methodologies.

Beyond methodological challenges, there are significant defects in training data. Notably, the sparsity issue in predicate representations poses challenges, similar to the long-tail effect, complicating the task of visual relation detection. Using the Stanford VRD dataset \cite{r16} as an illustrative case, its potential predefined relations count exceeds 700k, highlighting that numerous real relations lack any training samples. UVTransE \cite{r51} introduces joint features for subject and object boundaries, effectively capturing contextual information. This class of analogy transfer-based SSG methods exhibits promising applications, and empirical findings substantiate their effectiveness.

\section{details of dataset}
This section details the dataset, including its source and preprocessing, object and relation categories, and annotation methods.

\subsection{Images Sources and Dataset Splits}
To efficiently construct a large-scale remote sensing relation dataset annotated with OBB, we selected the FAIR1M dataset as our foundation. FAIR1M is a large-scale standard benchmark dataset used for fine-grained object detection and recognition tasks in high-resolution remote sensing images. It contains over one million instances and more than 40,000 images collected from high-resolution satellites and Google Earth, with resolutions ranging from 0.3 to 0.8 meters. FAIR1M employs OBB annotations, categorizing objects into 5 major classes and 37 fine-grained category. We used images from FAIR1M as our source, selecting 21,392 images for additional object bounding box and relation annotations.

For the dataset split, we used half of the images for the training set, one-sixth for the validation set, and one-third for the test set. The dataset includes both the original images and their corresponding OBB and relation annotations. Detailed split information for the objects and relations is shown in Table~\ref{tab:obj_category} and Table~\ref{tab:rel_category}.

\subsection{Category Design}

The original annotations of object categories in FAIR1M did not meet our requirements for constructing a large-scale remote sensing relation dataset. Therefore, while retaining the original FAIR1M annotations, we conducted extensive re-annotations to expand the object categories and added a substantial number of relation annotations.

\begin{table}[t!]
\centering
\caption{\\The number of instances for each relation category and dataset split.}
\label{tab:rel_category}
\begin{tblr}{
  colspec = {X[2.5,c] X[1,c] X[1,c] X[1,c]},
  rows = {m},
  row{1} = {font=\bfseries},
  hline{1-2, Z} = {solid},
  row{3-Z} = {rowsep=0.25pt}
}
Category & Train & Validation & Test \\
park at & 103368 & 34179 & 69634 \\
park next to & 98734 & 31830 & 65712 \\
close to & 79292 & 25575 & 48970 \\
accessible & 70097 & 22669 & 45514 \\
drive on & 32903 & 11014 & 20853 \\
moor & 24648 & 8397 & 14557 \\
serve & 11433 & 3657 & 7787 \\
parallel & 11102 & 3701 & 7074 \\
adjacent & 8036 & 3149 & 5567 \\
sail on & 6471 & 2109 & 4694 \\
belong to & 6617 & 2147 & 4461 \\
pile up & 5122 & 1834 & 3179 \\
inside & 98334 & 31644 & 61564 \\
cross & 4089 & 1410 & 2628 \\
supplement & 2331 & 633 & 1373 \\
supply & 1815 & 619 & 1114 \\
slow & 1649 & 594 & 1142 \\
contain & 1462 & 618 & 1058 \\
taxi on & 1528 & 516 & 1082 \\
cooperate & 1206 & 422 & 712 \\
power & 655 & 186 & 341 \\
link & 655 & 155 & 315 \\
preparation & 515 & 228 & 343 \\
above & 508 & 172 & 394 \\
hoist & 566 & 188 & 295 \\
under & 229 & 93 & 218 \\
ventilate & 248 & 89 & 161 \\
on & 216 & 73 & 183 \\
transport & 209 & 78 & 178 \\
construction & 277 & 54 & 98 \\
sail by & 186 & 22 & 102 \\
tow & 55 & 17 & 53 \\
block & 27 & 8 & 14 \\
connect & 189 & 34 & 101 \\
drive away from & 45 & 19 & 21 \\
enter & 133 & 41 & 71 \\
away from & 101 & 27 & 64 \\
dock alone at & 518 & 189 & 309 \\
park alone at & 213 & 89 & 134 \\
drive at the same lane & 1980 & 742 & 1351 \\
drive at the different lane & 1589 & 562 & 1086 \\
typhoon impact & 78 & 35 & 48 \\
load & 7 & 1 & 9 \\
pass under & 18 & 4 & 9 \\
intersect & 168 & 67 & 95 \\
around & 32 & 11 & 17 \\
emit & 9 & 3 & 6 \\
own & 21 & 7 & 18 \\
stick to & 987 & 359 & 571 \\
separate & 13 & 2 & 5 \\
transfer passenger & 76 & 25 & 39 \\
mirror symmetry & 9 & 4 & 7 \\
symmetry & 35 & 19 & 24 \\
converge & 451 & 132 & 279 \\
border & 128 & 52 & 76 \\
dock at & 295 & 98 & 186 \\
support & 84 & 37 & 51 \\
manage & 298 & 101 & 209 \\
shuttle & 68 & 31 & 40 \\
command & 31 & 8 & 15 \\
dig & 43 & 9 & 26 \\
cultivate & 7 & 2 & 9 \\
forest fire & 29 & 10 & 17 \\
pull & 15 & 14 & 12 \\
\end{tblr}
\end{table}

\subsubsection {Objects Category} 
FAIR1M includes 5 major categories and 37 fine-grained object categories, with the fine-grained categories containing richer semantic information. We chose to retain these 37 fine-grained categories as the true labels for the original bounding boxes. However, since these 37 categories only belong to 5 major classes, the relation information they contain is still not sufficiently rich. Hence, there is a need to expand the object categories. Each category could have semantic or spatial relations with others, so increasing the number of categories significantly enhances the richness of the relation information.
Building upon the original FAIR1M annotations and the 37 categories ( passenger ship, bridge, motorboat, fishing boat, tugboat,intersection,……), we added 23 new categories: crane, harbor, water, building, block, parking lot, road, solar panel, exhaust fan, airport, runway, gas station, pool, container, factory, expressway service area, storage tank, storage tank group, construction site, dam, locomotive, train carriage, cargo. These categories were selected by remote sensing image interpretation experts based on their prevalence, relation information, and practical value in applications. Typical objects corresponding to each category are illustrated in Fig.~\ref{fig:obj_sample}.

\subsubsection {Relation Category}

After in-depth discussions with professional remote sensing image interpretation experts, we identified 64 common and practically valuable relation categories in remote sensing scenarios. These can be divided into two major categories based on their focus: spatial relations and semantic relations. Spatial relations, which focus on the spatial distribution relations of objects, include 20 types: close to, park next to, adjacent, on, above, under, inside, moor, park at, cross, accessible, parallel, enter, around, stick to, mirror symmetry, symmetry, drive at the same lane, drive at the different lane, pass under. Semantic relations, on the other hand, emphasize advanced semantic relations between objects and include 44 types: hoist, cooperate, supply, tow, sail on, drive on, taxi on, sail by, transport, serve, belong to, contain, power, ventilate, pile up, supplement, slow, preparation, block, construction, pull, link, connect, drive away from, sail away from, dock alone at, park alone at, intersect, emit, own, separate, transfer passenger, converge, border, support, manage, command, shuttle, dig, cultivate, forest fire, typhoon impact, load. Typical diagrams for each type of relation are illustrated in Fig.~\ref{fig:rel_sample}.

\begin{figure*}[t!]
\centering
\includegraphics[width=7in]{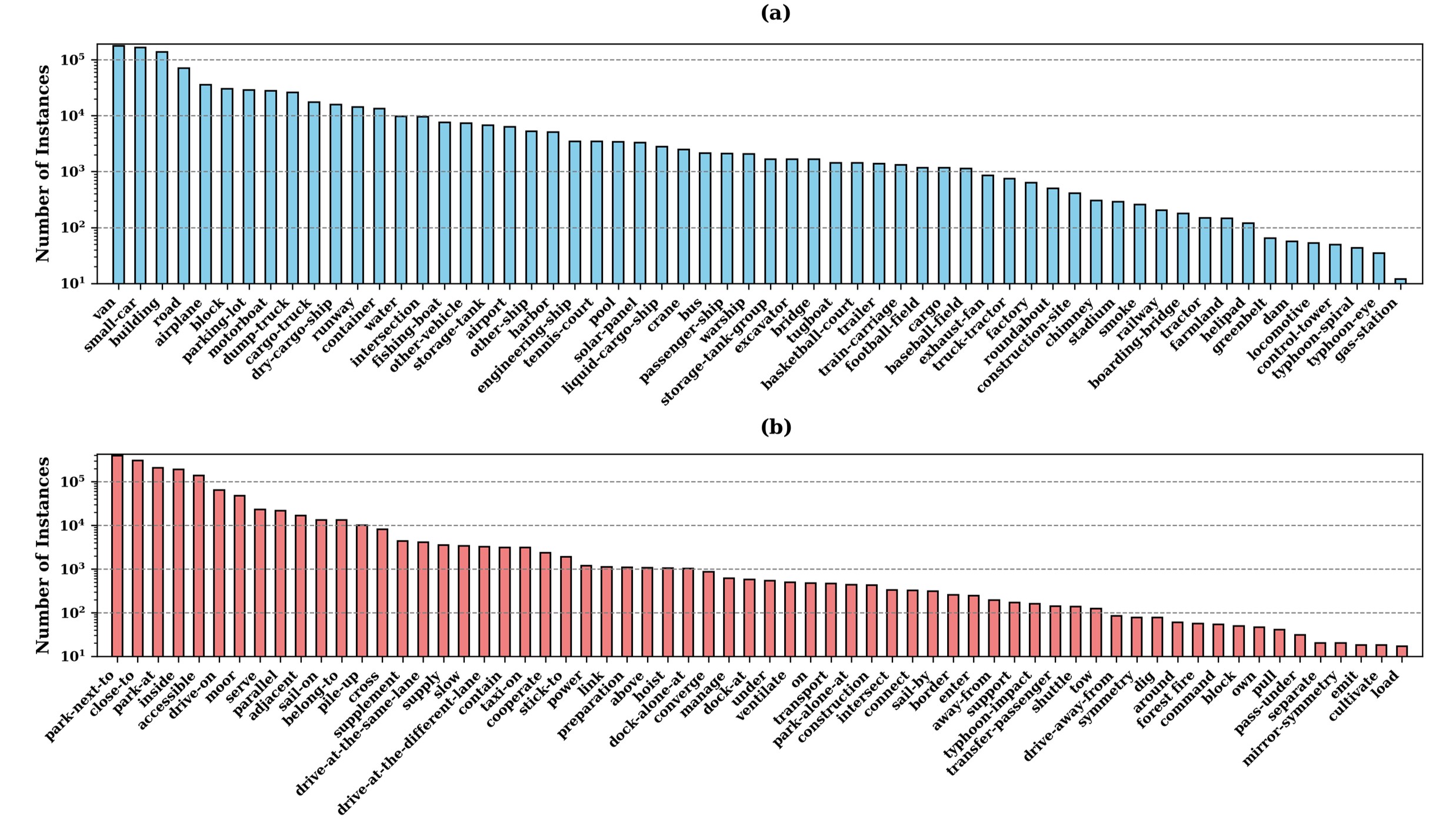}
\caption{(a)The distribution of the number of instances per object category, (b)The distribution of the number of instances per relation category.}
\label{fig:dist_of_2}
\end{figure*}

\subsection {Image Annotaion}
\subsubsection {Annotation Format}

Our relation dataset requires two types of annotations: object annotations and relation annotations. Common relation datasets in natural scenes, such as Visual Genome and HVCRD, use horizontal bounding boxes (HBB) without directional information for object annotations. This is because their images come from natural scenes, where the images are smaller, the objects are simpler, and there are fewer distracting elements. However, the span of remote sensing images is much broader than that of natural scenes, and their object complexity is significantly higher. Using HBB for annotations in remote sensing could compromise the precision of the annotations. For example, if a road diagonally crosses the entire image, using HBB would mean the size of the bounding box is the size of the entire image, with foreground pixels significantly outnumbered by background pixels. This leads to inaccurate object information extraction, thus affecting the accuracy of relational recognition between objects.

Using OBB $\{(x_i, y_i), i=1,2,3,4\}$, where $(x_i, y_i)$ represents the vertices of the bounding box, addresses these issues effectively. For objects in remote sensing images that have a clear forward orientation, such as airplanes, ships, and vehicles, we choose the vertex on the front left as the first vertex of the OBB and annotate the other three vertices clockwise. For objects without a clear forward orientation, such as oil tanks, containers, and pools, we select the top-left vertex as the first vertex and mark the remaining three vertices clockwise. For relation annotations, similar to most natural scene relation datasets, we use the \textless subject, relation, object\textgreater triplet format. Each subject and object is one of the previously mentioned 60 object categories, and each relation is one of the 64 relationa categories mentioned earlier.

\subsubsection{Annotation Procedure}

We develop a specialized annotation tool to add relation annotations between object bounding boxes, and also develope a dedicated visualization tool to display the relation annotation information for each image. To ensure the accuracy of the dataset annotations, we established a comprehensive annotation process for the remote sensing relation dataset:

\begin{itemize}
\item Discussion with Experts. We engage with professional remote sensing image interpreters to select object and relation categories that are common and have practical application value based on the existing dataset images and real-world scenarios.

\item Example Selection and Annotation Guidelines. Professionals select typical examples for each object and relation category from the images to be annotated. They provide specific guidelines and examples for each category to ensure consistency and accuracy.

\item Standardization of Annotation Manual. A standardized annotation manual is written, and professionals instruct the annotators on the standards and important considerations required for accurate annotations.

\item Dual Annotation Teams. The annotators are divided into two groups, A and B. Both groups concurrently perform object annotations and relation annotations. After completing the annotations, each group uses the visualization tool to check the accuracy of the other group's annotations.

\item Quality Control and Review. Approximately 20$\%$ of the annotated data is randomly selected and reviewed by professionals. If any issues are identified, feedback is provided to the annotators for correction, and this process is repeated until the professionals confirm the annotations are accurate.
\end{itemize}

This rigorous process ensures that our dataset not only meets high standards of accuracy but also is useful for developing and training models in remote sensing image interpretation and relation analysis.

\begin{figure*}[!t]
\centering
\includegraphics[width=7in]{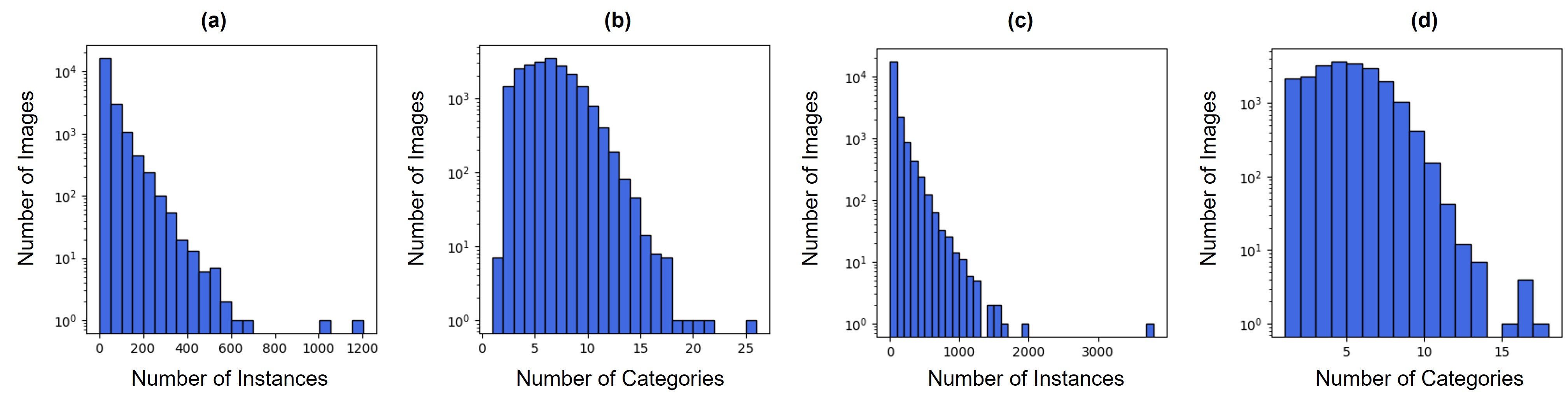}
\caption{(a) The distribution of the number of object instances per image. (b) The distribution of the number of object categories per image. (c) The distribution of the number of relation instances per image. (d) The distribution of the number of relation categories per image.}
\label{fig:4in1}
\end{figure*}

\begin{figure}[t!]
\centering
\includegraphics[width=0.85\linewidth]{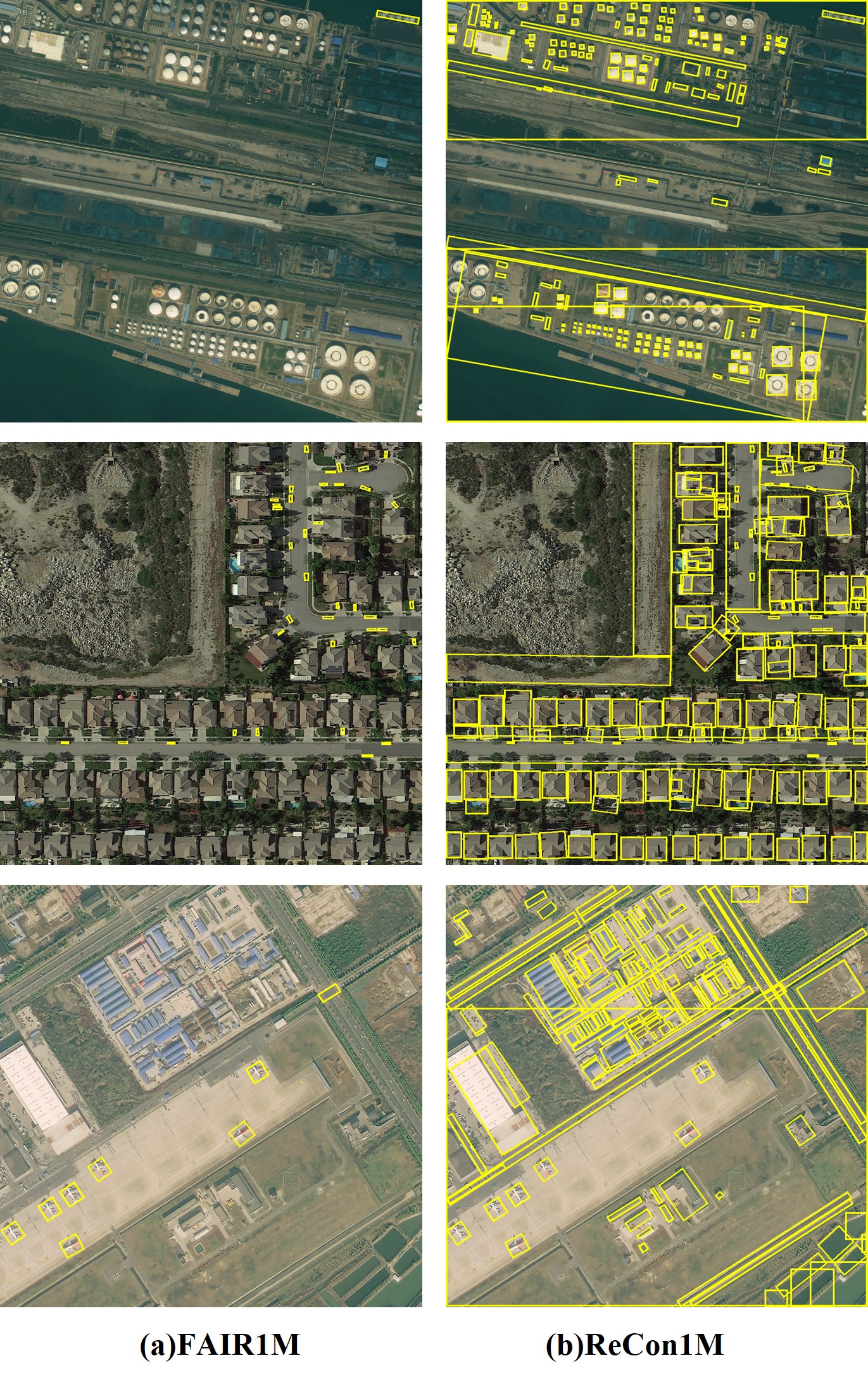}
\caption{Comparison of bounding box annotations in the same source images between FAIR1M and ReCon1M. (a) Annotations of FAIR1M, (b) Annotations of ReCon1M.}
\label{fig:compare_to_fair1m}
\end{figure}

\subsection{Dataset Statistics}
The input data and annotation formats for this dataset have already been described in Section III. Given the characteristics of this dataset, we will conduct two object detection tasks and one relation detection task on it. The object detection tasks include HBB detection and oriented bounding box detection. After performing object detection, several object bounding boxes and their categories will be obtained. In the relation detection task, for a given input image, we will output several relational triplet instances \textless subject, relation, object\textgreater, where both subject and object include the coordinates and categories of the bounding boxes, and relation represents the association description between subject and object.
\subsubsection{Instance}
ReCon1M includes 859,751 object instances with category annotations. We have added a large number of dense object annotations based on FAIR1M. Fig.~\ref{fig:compare_to_fair1m} illustrates the difference in the number of object annotations between FAIR1M and ReCon1M on the same image.As shown in Table~\ref{tab:obj_category}, there is a significant disparity in the number of instances across different categories. For example, the "van" category has the highest number of instances at 175,371, accounting for 20.3975$\%$ of total object instances. In contrast, the "solar panel" category, which is at the median, has 3,276 instances, making up 0.3810$\%$ of the total. The "gas station" category has the fewest instances at only 12, representing only 0.0014$\%$ of the total. This is a typical example of class imbalance in remote sensing image datasets, where some categories far outnumber others. This imbalance reflects real-world remote sensing scenarios and adds to the challenges of our dataset. Fig.~\ref{fig:4in1} also shows the distribution of the number of instances per image, with up to 1,200 instances in some images and an average of 40 per image, which is a significant increase compared to previous SGG datasets.

\begin{figure*}[t!]
\centering
\includegraphics[width=\linewidth]{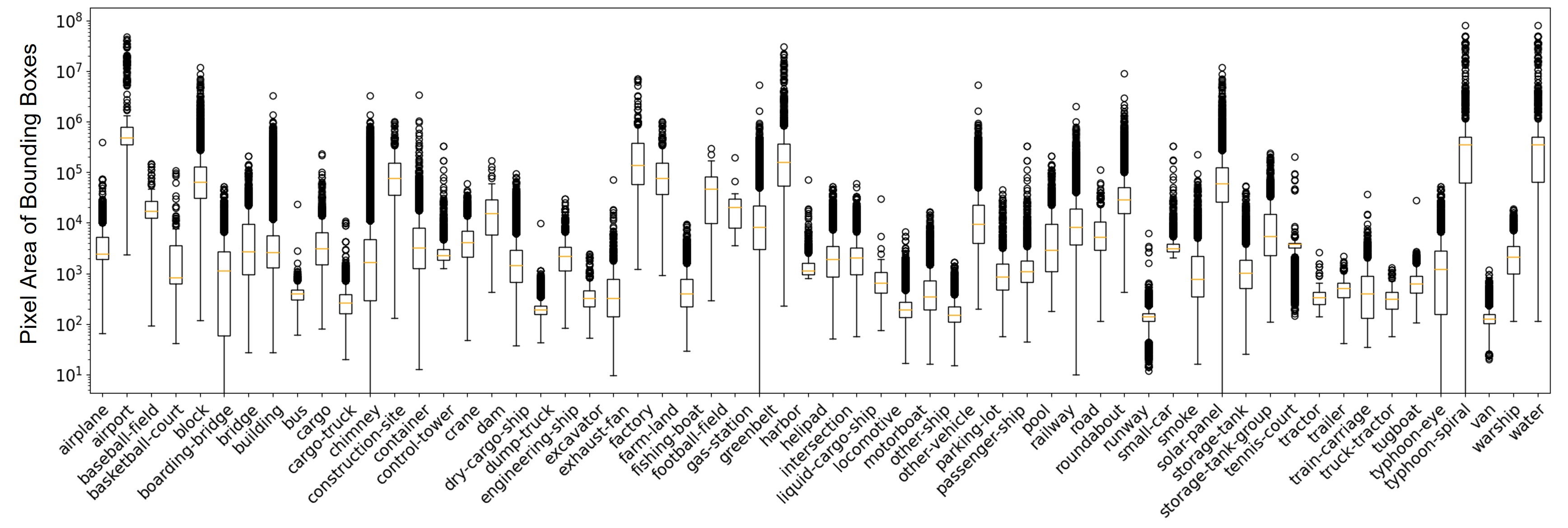}
\caption{Size variations for each object category in ReCon1M. The sizes of different categories vary in different ranges.}
\label{fig:box_plot}
\end{figure*}

\begin{figure}[t!]
\centering
\includegraphics[width=\linewidth]{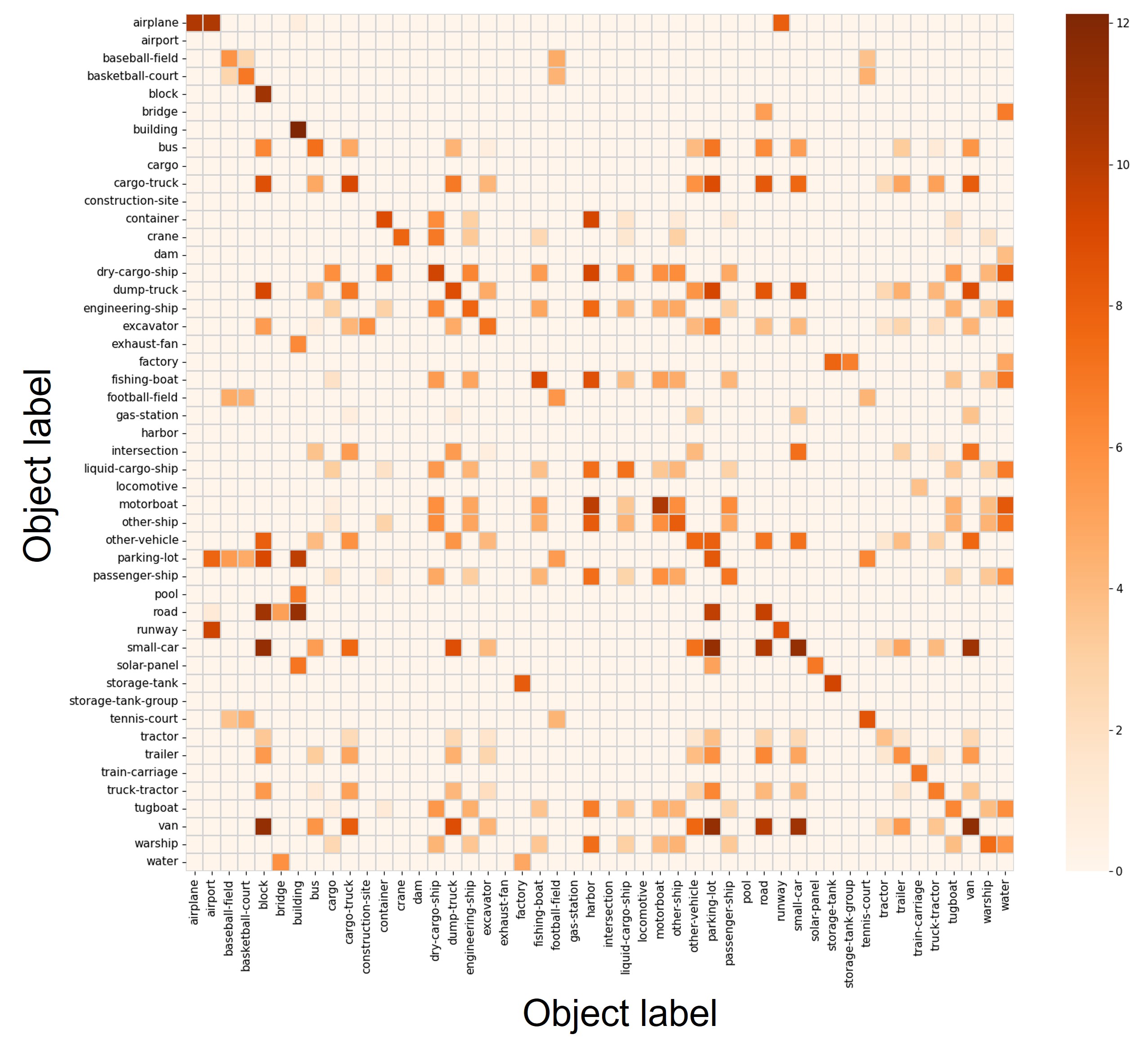}
\caption{Diagram of correlation strength between some object categories. Each grid point's value represents the logarithm of the number of relations occurring between two object categories. The darker the color, the higher the correlation strength between the two object categories, and vice versa.}
\label{fig:heatmap}
\end{figure}

Remote sensing images cover a wide area with a large variance in object scale sizes, posing significant challenges for object detection and relation recognition. We classify the objects into three sizes based on the area (bounding box size): large objects with pixel counts of 2048 or more, medium objects with pixel counts between 144 and 2047, and small objects with pixel counts between 11 and 143. 
The statistical results show that the proportions of large, medium, and small objects are 34.8\%, 38.6\%, and 26.6\% of the total number of objects, respectively.The relation between object scale and category is also significant, as shown in Fig.~\ref{fig:box_plot}. For example, the average area of category 'parking lot' is 22095.2, whereas for category 'van', it is only 132.7. There are also large variations in size within the same category; for example, the largest area in category 'building' is 3,315,169, and the smallest is 27.
\subsubsection{Relation}
Relations are the core component of our dataset, represented as relation triplets \textless subject, relation, object\textgreater, which we refer to as relation instances. Our dataset contains 1,149,342 relational annotations. Fig.~\ref{fig:dist_of_2} shows the number of relation instances for each relation category. Clearly, the distribution of relations exhibits a typical long-tail pattern. The most frequent relation category is "park at", with 207,181 instances, accounting for 18.0260$\%$ of all relation instances, while the least frequent is "pull", with only 41 instances, or 0.0036$\%$ of the total. Like the object categories, there is significant imbalance among the relation categories. 
Additionally, the distribution of relations is biased and often exhibits a strong correlation with the object categories, as demonstrated in Figure 8. The bias in the dataset reflects real-world laws, as, for example, boats are mostly seen sailing on water rather than cars, and airplanes taxi on runways rather than ships. This data bias is a critical issue in real-world applications, lending significant relevance and added challenge to our dataset.

\section{ALGORITHM ANALYSIS}
\subsection{Object Detection}
\subsubsection {Evaluation Tasks and Metrics}

Current methods in Scene Graph Generation (SGG) are mainly divided into one-stage and two-stage approaches, with the latter being more established and numerous. Object detection is a crucial component of the two-stage approaches in SGG. The RR dataset, with its extensive annotations of directed bounding boxes, allows for object detection tasks to be conducted. Models trained on this dataset can serve both as detector components for SGG tasks and be independently evaluated for their object detection capabilities on the RR dataset. Object detection tasks primarily include two types: HBB detection and OBB detection. HBB detection, which uses axis-aligned rectangular boxes, is suitable for objects with regular shapes or where orientation is not crucial, offering simplicity and computational efficiency. In contrast, OBB detection utilizes rotated rectangular boxes to locate objects, which is more suitable for irregularly shaped or direction-specific objects, such as ships or airplanes, providing more precise position and shape information, but with a higher computational complexity. Although our dataset only has annotations for OBB, since OBB annotations can be directly converted into HBB annotations, we conduct both types of object detection tasks for benchmark testing on this dataset.

We use mean average precision (mAP), a widely applied metric in object detection, as our evaluation standard. Precision refers to the proportion of actual objects among those detected by the model, calculated as the number of true positives (TP) divided by the sum of TP and false positives (FP):
\begin{align}
Precision=\frac{TP}{TP+FP}
\end{align}
For each object category, the average precision (AP) measures the area under the precision-recall curve at different confidence scores. Typically, detection results are sorted by confidence thresholds, and precision and recall are calculated at each threshold to construct the precision-recall curve. The AP represents the area under this curve. The mAP is the mean of the AP across all object categories, providing a comprehensive evaluation that reflects performance across multiple categories, not just a single one.

\begin{table*}[!t]
\centering
\caption{\\Baseline Results of Object Detection on ReCon1M}
\label{tab:det_results}
\begin{tblr}{
  colspec = {c X[c] X[c] X[c] X[c] X[c] X[c] X[c] X[c]},
  rows = {m},  % 设置所有行的单元格内容垂直居中
  row{1} = {font=\bfseries},
  cell{1}{2} = {c=4}{c,font=\bfseries},
  cell{1}{6} = {c=4}{c,font=\bfseries},
  hline{1-3, 24, Z} = {solid}, 
  vline{2, 6} = {2-Z}{solid},
  row{3-Z} = {rowsep=0.5pt}
}
    & HBB Results  &  &  &  & OBB Results  &  &  &  \\
Category & Faster R-CNN\cite{r5} & Cascade R-CNN\cite{cai2018cascade} & FCOS\cite{tian2020fcos} & RetinaNet\cite{lin2017focal} & Rotated Faster R-CNN\cite{r5} & Rotated RetinaNet\cite{lin2017focal} & ROI Transformer\cite{ding2019learning} & ReDet\cite{han2021redet} \\
van                & 0.189 & 0.247 & 0.098 & 0.153 & 0.475 & 0.333 & 0.514 & 0.570 \\
small car          & 0.199 & 0.252 & 0.130 & 0.181 & 0.500 & 0.399 & 0.535 & 0.591 \\
building           & 0.097 & 0.135 & 0.128 & 0.139 & 0.289 & 0.250 & 0.300 & 0.313 \\
road               & 0.076 & 0.119 & 0.053 & 0.047 & 0.164 & 0.055 & 0.208 & 0.243 \\
airplane           & 0.802 & 0.826 & 0.765 & 0.724 & 0.908 & 0.908 & 0.908 & 0.908 \\
block              & 0.200 & 0.239 & 0.215 & 0.187 & 0.398 & 0.385 & 0.407 & 0.416 \\
parking lot        & 0.103 & 0.129 & 0.133 & 0.136 & 0.297 & 0.292 & 0.319 & 0.326 \\
motorboat          & 0.359 & 0.445 & 0.207 & 0.250 & 0.616 & 0.287 & 0.647 & 0.698 \\
dump truck         & 0.206 & 0.309 & 0.025 & 0.034 & 0.421 & 0.087 & 0.489 & 0.501 \\
cargo truck        & 0.301 & 0.367 & 0.136 & 0.158 & 0.459 & 0.179 & 0.510 & 0.514 \\
dry cargo ship     & 0.459 & 0.541 & 0.307 & 0.251 & 0.571 & 0.195 & 0.665 & 0.687 \\
runway             & 0.113 & 0.182 & 0.092 & 0.096 & 0.218 & 0.052 & 0.291 & 0.325 \\
container          & 0.100 & 0.127 & 0.104 & 0.103 & 0.199 & 0.140 & 0.259 & 0.278 \\
water              & 0.327 & 0.366 & 0.340 & 0.365 & 0.433 & 0.431 & 0.481 & 0.491 \\
intersection       & 0.346 & 0.351 & 0.278 & 0.239 & 0.619 & 0.595 & 0.634 & 0.696 \\
fishing boat       & 0.229 & 0.324 & 0.097 & 0.052 & 0.257 & 0.109 & 0.328 & 0.405 \\
other vehicle      & 0.038 & 0.076 & 0.001 & 0.002 & 0.028 & 0.023 & 0.039 & 0.045 \\
storage tank       & 0.095 & 0.119 & 0.126 & 0.156 & 0.338 & 0.347 & 0.370 & 0.391 \\
airport            & 0.539 & 0.622 & 0.567 & 0.333 & 0.581 & 0.623 & 0.607 & 0.609 \\
...                & ...   & ...   & ...   & ...   & ...   & ...   & ...   & ...   \\
gas-station        & 0.000 & 0.000 & 0.000 & 0.000 & 0.000 & 0.000 & 0.000 & 0.000 \\
mAP                & 0.238 & \textbf{0.281} & 0.164 & 0.154 & 0.341 & 0.219 & 0.380 & \textbf{0.401} \\  
\end{tblr}
\end{table*}

\begin{figure*}[!t]
\centering
\includegraphics[width=\linewidth]{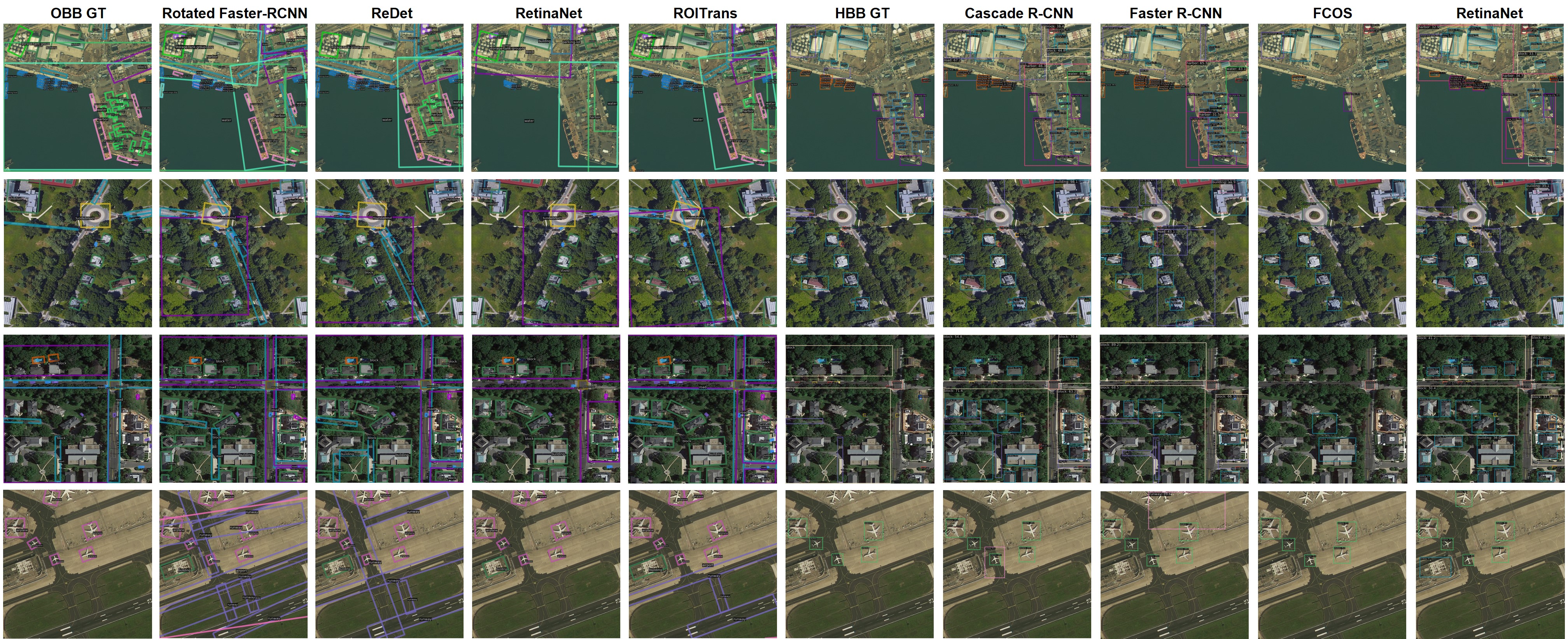}
\caption{The visualization of detection results.}
\label{fig:detection_visualization}
\end{figure*}

\subsubsection {Baseline Models}

For the two detection tasks, we selected four representative models for evaluation. In the HBB detection task, we chose Faster R-CNN, Cascade R-CNN, CenterNet, and RetinaNet as baselines. For the oriented bounding box detection task, we selected Rotated Faster R-CNN, Rotated RetinaNet, RoI Transformer, and ReDet as baselines.
\subsubsection {Training Details}

Our dataset contains very-high-resolution images (e.g., 6000×6000), which most baseline models cannot process directly as input. Therefore, we first divided the images into 800×800 blocks with a stride of 400. For both horizontal and oriented box detection tasks, we used standard hyperparameters from models in mmdetection and mmrotate. Each model was trained on two RTX 3090 GPUs, with Adam as the optimizer and an initial learning rate of 0.005. A stepwise learning rate schedule was employed, and training was conducted for a total of 24 epochs.

\subsubsection {Results and Analysis}
In our study, we conducted eight distinct experiments using varied model frameworks, the results of which are shown in Table~\ref{tab:det_results}. This exploration reveals that the choice of model architecture plays a pivotal role in influencing the mAP metric. Specifically, in the HBB task, the Cascade R-CNN model tops the chart with an mAP of 0.281, contrasting sharply with RetinaNet's lowest score of 0.154. In the Oriented Bounding Box (OBB) task, ReDet excels with an mAP of 0.401, markedly outperforming the Rotated RetinaNet, which scored the lowest at 0.219. Moreover, the substantial variance in AP scores across different classes, exacerbated by class imbalance, is characteristic of remote sensing imagery. Notably, in both the HBB and OBB tasks, aircraft objects demonstrate the highest AP scores, significantly surpassing other categories. This phenomenon likely results from their ample annotations and distinct textural and geometric features, which facilitate the detector's training on more discriminative characteristics, thereby boosting accuracy. Conversely, the least annotated categories in the dataset (dams, locomotives, and gas stations), which have fewer than 100 annotations each, suffer from insufficient training data. This lack of data prevents effective learning of discriminative features, resulting in zero average precision (AP) scores across all experiments.

Additionally, our findings indicate that the mAP scores for OBB tasks notably exceed those for HBB tasks. This is intriguing, considering that OBB tasks are generally deemed more complex than HBB tasks due to the additional requirement of predicting the orientation of the bounding boxes. From the data in Table~\ref{tab:det_results}, the highest mAP recorded in OBB experiments was 0.401, significantly surpassing the highest HBB score of 0.281. This superior performance in OBB could be attributed to several factors: 1) Dense annotations in the dataset may reduce overlap between adjacent objects, which is particularly beneficial for slanted boxes that can more accurately distinguish adjacent objects. 2) OBB annotations during training and detection provide richer and more precise feature representations, aiding the model in better recognizing objects amidst complex backgrounds and diverse orientations. 3) Horizontal annotations tend to include irrelevant background elements when handling inclined objects, which can deform features and impair detection accuracy. In contrast, slanted annotations adapt better to the actual contours and orientations of objects, mitigating the detrimental effects of deformation.

\begin{figure*}[!t]
\centering
\includegraphics[width=\linewidth]{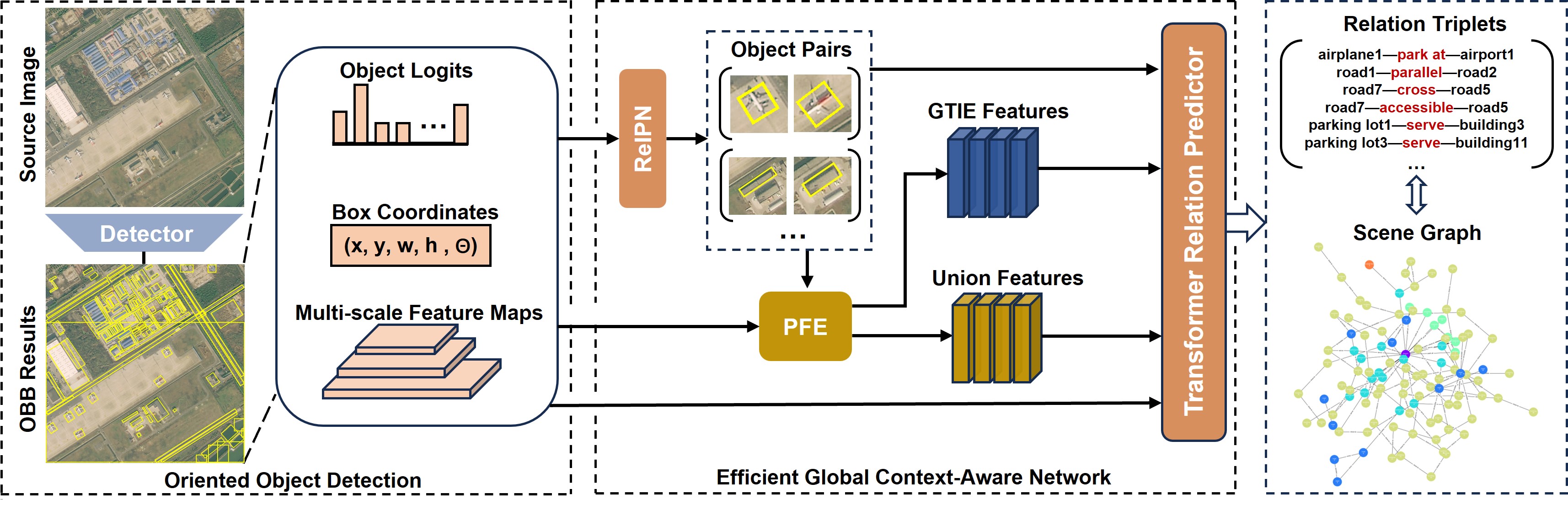}
\caption{The overall structure of EGCA-Net. RelPN refers to related object pair proposal, PFE refers to pair-wise feature extraction, and GTIE refers to geometric topological information embedding.}
\label{fig:model_structure}
\end{figure*}

\subsection{Scene Graph Generation}
\subsubsection {Evaluation Tasks and Metrics}

SGG focuses on identifying objects in images and determining the relations between them. To comprehensively assess the capabilities of our dataset and the effectiveness of various algorithms, we have segmented the SGG task into three distinct sub-tasks: Predicate Classification (PreCls), Scene Graph Classification (SGCLS), and Scene Graph Detection (SGDET). PreCls involves predicting the relations between pairs of objects given their labels and bounding boxes. The primary challenge here is to accurately classify the type of relation without misinterpreting the visual and spatial cues. In SGCLS, both the object labels and their relations need to be classified, assuming that the bounding boxes of objects are known. This tests the algorithm’s ability to integrate object recognition with relation prediction. SGDET is the most comprehensive sub-task, where the algorithm must detect objects, classify them, and predict their interrelations, starting from raw images. It simulates a real-world scenario where both objects and their relations need to be inferred without prior knowledge. The current methods in SGG can also be divided into one-stage and two-stage methods. Compared with the vigorous development of two-stage methods, one-stage methods are still in the initial stage. Two-stage methods typically achieve higher accuracy in both object detection and relation prediction compared to one-stage methods. Additionally, many existing benchmark datasets and evaluation metrics in the field of SGG are designed with two-stage methods in mind. Primarily focusing on two-stage methods can facilitate a more meaningful evaluation and comparison of results.

In the relation detection task, we have opted to use Mean Recall at K (mR@K) as the evaluation metric for model performance. Recall is defined as the proportion of actual relational triplets successfully detected by the model out of the total real relational triplets. It is calculated as the number of True Positives (TP) divided by the sum of TP and False Negatives (FN):
\begin{align}
Recall=\frac{TP}{TP+FN}
\end{align}
Recall@K is an extended metric of recall, where K represents the number of top highest-scoring predictions retained for calculating Recall, focusing specifically on the successful detection of real relational triplets within these predictions. This approach allows researchers to select different K values based on varying performance assessment needs. In experiments with natural scene datasets, the K value is typically set to 20, 50, or 100. However, due to the significantly higher annotation density of ReCon1M compared to natural scene datasets, we additionally set the K value to 500. mR@K is the mean of the Recall@K values across various relational categories.

\subsubsection {Baseline Models}
We have selected ten two-stage models in SGG for evaluation, including IMP, Neural MOTIF, RTN, VCTree, MSDN, RU-Net, HL-Net, Seq2Seq, BGNN, SQUAT and the proposed EGCA-Net. These models include both classic and the latest state-of-the-art (SOTA) models, which can effectively serve as benchmarks for the SGG task on our dataset.

\subsubsection {Proposed Method}
Since the baseline methods we adopted are designed for relation prediction tasks based on HBB in natural scenes, while ReCon1M is a dataset annotated with OBB and used in remote sensing scenes, directly replacing the detector in the baseline methods with a rotated detector does not yield satisfactory results. Therefore, to address the issues of OBB annotations, dense objects and relations, and large variations in object scales and distances between objects in ReCon1M, we improved on previous work \cite{r58} and proposed a Efficient Global Context-Aware Network (EGCA-Net). The overall structure is shown in Fig.~\ref{fig:model_structure}. It mainly consists of four stages: oriented object detection, related object pair proposal, pair-wise feature extraction, and relation triplets prediction.

\begin{itemize}
  \vspace{-1mm}
    \item 
        \textbf{Oriented Object Detection}. Considering that most two-stage scene graph generation models previously used Faster R-CNN as the object detector, we adopt Rotated Faster R-CNN as the OBB detector. The main difference between Rotated Faster R-CNN and Faster R-CNN is that the former can detect objects at any angle of rotation. It achieves this through an improved Region Proposal Network (RPN), using rotated candidate regions and a rotated RoI pooling method, and adding an angle parameter in the bounding box regression, making it more suitable for Oriented object detection tasks. In two-stage SGG methods, the relation prediction part relies on the accuracy of the detector and the quality of the extracted visual features. Using Rotated Faster R-CNN, specifically designed for OBB detection, can better apply to the ReCon1M dataset and better support the subsequent relation prediction components. In this stage, the detector will output the encoded coordinates of \(N\) bounding boxes \(D_{i}=\left\{\left(x_{i}, y_{i}, w_{i}, h_{i}, \theta_{i}\right), i=1,2, \ldots, N\right\}\), multi-scale feature maps \(F_{fpn}\), and the logits for each box \(l_i=\begin{bmatrix}l_{i,1},l_{i,2},...,l_{i,K}\end{bmatrix},i\in\{1,2,...,N\}\) where K refers to the total number of object categories.
    
\begin{table*}[!t]
\centering
\caption{\\Baseline Results of Scene Graph Generation on ReCon1M}
\label{tab:sgg_results}
\begin{tblr}{
  % width = 0.95\textwidth,
  colspec = {c X[c] X[c] X[c] X[c] X[c] X[c] X[c] X[c] X[c] X[c] X[c] X[c]},
  rows = {m},  % 设置所有行的单元格内容垂直居中
  row{1} = {font=\bfseries},
  cell{1}{1} = {c=13}{c},
  cell{2}{2} = {c=4}{c},
  cell{2}{6} = {c=4}{c},
  cell{2}{10} = {c=4}{c},
  row{14} = {font=\bfseries},
  cell{14}{1} = {c=13}{c},
  cell{15}{2} = {c=4}{c},
  cell{15}{6} = {c=4}{c},
  cell{15}{10} = {c=4}{c},
  hline{1, 2, 3, 4, 14, 15, 16, 17, Z} = {solid}, 
  vline{2, 6, 10} = {2-14}{solid},
  vline{2, 6, 10} = {15-Z}{solid},
  % row{3-Z} = {rowsep=1pt},
  % row{2} = {font=\fontsize{7pt}{12pt}\selectfont},
  % row{3-Z} = {2-Z}{font=\fontsize{7pt}{12pt}\selectfont}
}
 Recall@K Results (K=20, 50, 100, 500)& & & & & & & & & & & & \\
 & PredCLS & & & & SGCLS & & & & SGDET & & & \\
Model & R@20 & R@50 & R@100 & R@500 & R@20 & R@50 & R@100 & R@500 & R@20 & R@50 & R@100 & R@500 \\
IMP\cite{r35}   &  59.8 &  75.4 & 83.9 &  92.0 & 52.3 & 64.1 &  71.3 & 76.7 & 17.2 & 24.1 & 30.5 & 45.8 \\
Motifs\cite{r29}  &63.1& 79.1 &  87.7 & 95.9 & 50.8 & 63.2 & 70.4 & 74.8 & 31.2 & 39.3 & 44.7 & 52.0 \\
RTN\cite{r58}  & \textbf{63.8} & 79.5 & 87.9 & 96.5 & \textbf{52.7} & \textbf{65.7} & \textbf{72.2} & \textbf{77.8} & 31.4 & 39.1 & 44.2 & 51.0 \\
VCTree\cite{tang2019learning}  &  63.2 & 79.1 & 87.8 & 96.0 & 50.6 & 62.0 & 71.2 & 74.7 &  31.4 & 39.6 & 44.9 & 52.0 \\
BGNN\cite{li2021bipartite}       & 59.3 & 75.3 & 84.7 & 95.0 & 49.7 & 61.7 & 68.4 & 74.9 & 31.8 & 39.9 & \textbf{45.5} & \textbf{53.2} \\
GPS-Net\cite{lin2020gps}      & 59.7 & 75.8 & 85.2 & 94.4 & 48.1 & 60.3 & 67.4 & 74.9 & 31.5 & 39.5 & 44.9 & 52.8\\
PE-Net\cite{zheng2023prototype} & 61.6 & 78.0 & 87.4 & 96.5 & 51.5 & 64.0 & 70.6 & 76.7 & 31.6 & 39.7 & 45.3 & 53.1\\
G R-CNN~\cite{yang2018graphrcnn}          & 61.5 & 77.9 & 87.0 & 96.0 & 49.0 & 61.4 & 68.4 & 75.6 & 31.1 & 38.7 & 44.0 & 52.0\\
SQUAT~\cite{jung2023devil}         & 60.9 & 77.0 & 85.9 & 95.7 & 48.5 & 60.4 & 67.5 & 75.3 & 31.5 & 39.6 & 45.3 & 52.9\\
\textbf{EGCA-Net}      & 63.5 & \textbf{79.8} & \textbf{88.1} & \textbf{96.6} & 51.9 & 63.6 & 70.9 & 77.4 & \textbf{31.9} & \textbf{40.2} & 45.1 & 53.1\\
 Mean Recall@K Results (K=20, 50, 100, 500)& & & & & & & & & & & & \\
 & PredCLS & & & & SGCLS & & & & SGDET & & & \\
Model & mR@20 & mR@50 & mR@100 & mR@500 & mR@20 & mR@50 & mR@100 & mR@500 & mR@20 & mR@50 & mR@100 & mR@500 \\
IMP  & 27.4 & 40.8 & 51.1 & 64.1 & 26.9 & 39.1 & 47.8 & 57.5 & 6.7 & 10.8 & 15.8 & 31.3 \\
Motifs & 38.4 & 55.1 & 66.5 & 82.7 & 27.4 & 38.1 & 46.4 & 56.0 & 13.5 & 19.9 & 24.9 & 34.4 \\
RTN   & 43.6 & 60.4 & \textbf{72.9} & 87.9 & \textbf{31.0} & \textbf{42.4} & 50.7 & 60.5 & 14.4 & 20.4 & 25.2 & 34.4 \\
VCTree  & 41.0 & 56.9 & 68.8 & 85.0 & 26.5 & 38.5 & 45.4 & 54.8 &  13.8 & \textbf{20.5} & 25.3 & 34.2 \\
BGNN  & 34.8 & 51.0 & 61.9 & 76.9 & 25.1 & 35.6 & 42.6 & 51.7 & 13.8 & 19.5 & 24.1 & 33.3 \\
GPS-Net   & 27.9 & 40.3 & 50.3 & 64.0 & 21.6 & 30.8 & 37.4 & 47.0 & 12.3 & 17.3 & 21.1 & 29.2\\
PE-Net  & \textbf{45.7} & \textbf{60.9} & 72.2 & 88.7 & 26.4 & 38.4 & 45.8 & 56.3 & 14.1 & 19.3 & 25.1 & 33.9\\
G R-CNN  & 35.3 & 50.7 & 61.7 & 78.2 & 24.2 & 34.4 & 41.4 & 51.8 & 12.5 & 17.1 & 21.3 & 28.7\\
SQUAT & 37.4 &53.1 & 64.8& 81.2 & 24.8 & 35.0 & 41.9 & 51.7 & 14.4 & 19.6 & 25.1 & 34.1\\
\textbf{EGCA-Net}  & 43.1 & 60.8 & \textbf{72.9} & \textbf{89.1} & 29.5 & 42.1 & \textbf{51.1} & \textbf{60.8} & \textbf{14.6} & 20.2 & \textbf{25.7} & \textbf{35.0}\\
\end{tblr}
\end{table*}

\begin{figure*}[h]
\centering
\includegraphics[width=\linewidth]{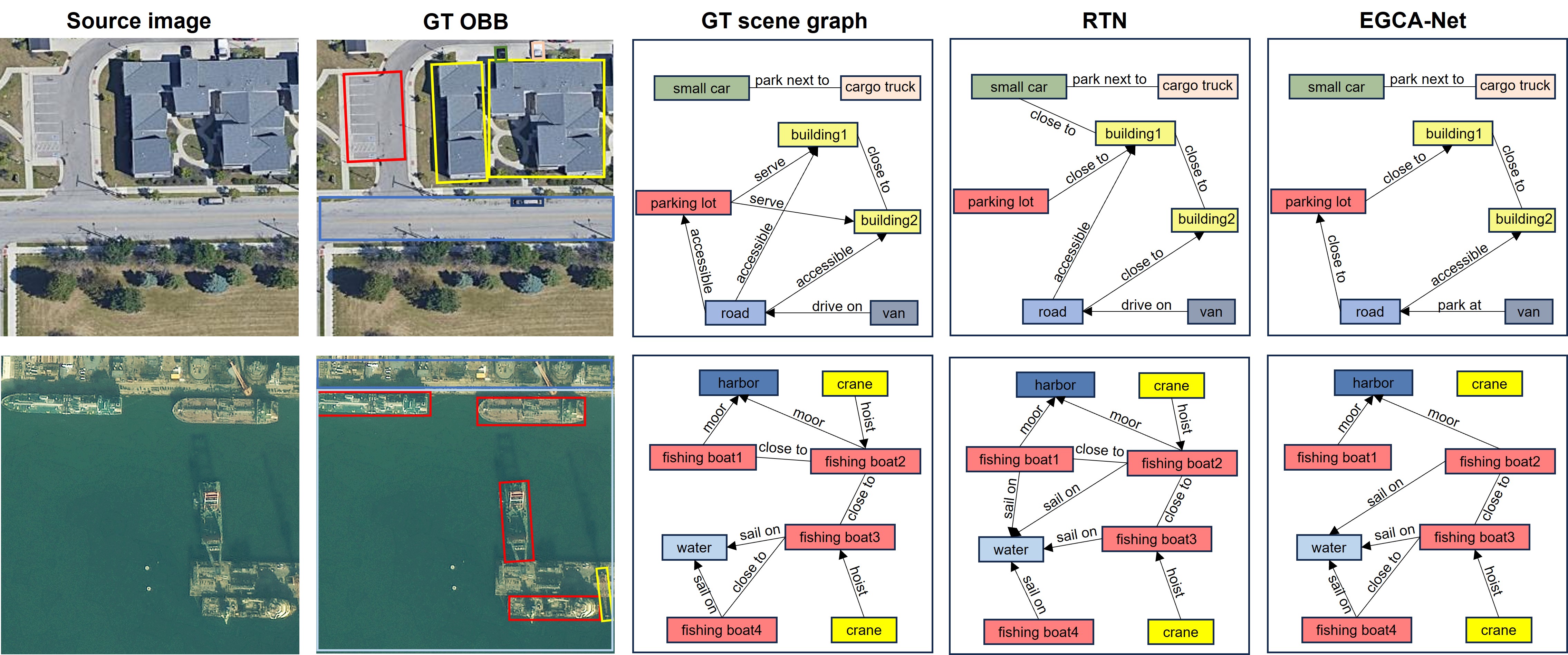}
\caption{Visualization Results of two baseline models(RTN and EGCA-Net). The two sets of visualizations from left to right are the source image, the image annotated with OBB, the ground truth (GT) scene graph, the Transformer-predicted scene graph, and the Motifs-predicted scene graph.}
\label{fig:sgg_exp_visulization}
\end{figure*}
    
    \item 
        \textbf{Related object pair proposal}. Theoretically, there are \(N*(N-1)\) possible potential relations between N objects, and one of the biggest differences between ReCon1M and previous natural scenes is that the object and relation annotations are much denser. As the number of objects N increases, the number of potentially related object pairs grows exponentially. During model training, we can solve this problem by sampling no more than the maximum specified number of positive and negative samples. However, during model inference testing, especially in SGDET tasks, it is necessary to predict the relations between all possible related object pairs output by the object detector, which means the demand for computing resources and the time taken for model inference both grow exponentially with the number of objects. This is unacceptable. Therefore, we propose a module called Related Object Pair Proposal Network(RelPN), which combines feature fusion and a multi-head self-attention mechanism. First, obtain the class of each object \(C_{i}=\arg\max(l_{i})\), from the logits of each object, then acquire the visual features \(F_{box}\) of each bounding box through ROI pooling. Next, concatenate \(C\), \(F_{box}\), \(D\)to obtain a fused feature matrix, and input this matrix into a multi-head attention mechanism module to get a logits matrix. Finally, calculate the binary cross-entropy loss between this logits matrix and the ground truth relation label matrix:
        \begin{align}
        L_{RelPN}=BCEL(l,R)
        \end{align}
      \item 
       \textbf{Pair-wise feature extraction}. The ultimate goal of the model is to predict possible related object pairs and the potential relations between them. Therefore, extracting more discriminative features at the pair-wise level is crucial. First, through ROI pooling, we can directly extract the feature map corresponding to each box from the multi-scale feature maps \(F_{fpn}\) output by the detector. However, the features obtained this way are limited, as they are only object-wise features. Therefore, for each object pair, we can extract additional object-wise visual features to aid in relation prediction. For the two bounding boxes in an object pair, their respective ROI features \(F_{subject}\) and \(F_{object}\) can be directly obtained by ROI pooling. Based on their coordinates, we can determine the coordinates of the smallest bounding rectangle which completely encloses these two boxes and use these coordinates to acquire the ROI feature \(F_{rectangle}\), which provides richer global context information through ROI pooling.For the three sets of features of an object pair, we concatenate them and then pass them through a $1 \times 1$ convolution layer to get fully fused features \(F_{union}\), retaining all the feature map information to enhance their global context awareness. Additionally, there are some geometric and topological information between the two objects in each object pair that can aid in relation prediction. For example, the distance between the centers of the two objects, the ratio of the areas between the two objects, the aspect ratio of the bounding box, the intersection over union (IoU) between the two bounding boxes, etc. We can concatenate these calculated feature values with the coordinates of the two objects to obtain an additional feature vector, which is then mapped to a higher-dimensional feature space through a fully connected layer to obtain \(F_{g}\). The feature vector embedded with pair-wise object geometric and topological information can assist subsequent relation prediction.
       \item 
        \textbf{relation triplets prediction}.Relation triplets prediction. We use a transformer as the relation predictor. We input bounding boxes's coordinates \(C\), object logits \(l\), and the corresponding ROI features \(F_{box}\) into the predictor to generate the contextual embedding \(E_{box}\) for each object and refine the object logits \(l_{refine}\) for the bounding boxes's categories (used for SGCLS and SGDET tasks). The object logits will be calculated with the true labels of each object \(C_{label}\) using cross-entropy loss to obtain \(L_{refine}\). Since objects have different feature representations when they are subjects and objects, we decode \(E_{box}\) through a linear layer into two different embeddings, \(E_{subject}\) and \(E_{object}\). Then, we concatenate the previously obtained geometric feature \(F_{g}\) with \(E_{subject}\) and \(E_{object}\), and combine them with the previously obtained union feature \(F_{union}\) to predict the relations for each object pair, resulting in the final relation triplets. Then, calculate the cross-entropy loss between the predicted relation triplets \(T_{pred}\) and the ground truth relation labels \(T_{label}\)  to obtain \(L_{relation}\)
        \begin{align}
        L_{refine} &= CEL(l_{refine}, C_{label}) \\
        L_{relation} &= CEL(T_{pred}, T_{label})
        \end{align}
        
       \item
        \textbf{Loss Function}. The entire framework goes through four stages: orented object detection, related object pair proposal, pair-wise feature extraction, and relation triplets prediction. The final loss is a weighted sum of the RelPN loss, object refine loss, and relation prediction loss:
        \begin{align}
        L_{CTECNet}=L_{RelPN}+L_{refine}+L_{relation}
        \end{align}

\end{itemize}

\subsubsection {Training Details}
Due to the two-stage approach of SGG methods, which separates object detection and relation prediction into two parts, we need to use a separately trained object detection model as a detector. For the sake of convenience in comparison with previous research, we have chosen to use the widely adopted Rotated Faster R-CNN as the detector. The image input settings for the detector are the same as those in the object detection experiments. When training the SGG model, we need to first load the weights of the previously trained detector and freeze the detector's parameters, training and updating only the parameters of the relation detector. For each task of each model, we train on a single RTX3090, using Adam as the optimizer with an initial learning rate of 0.0001 and employing cosine annealing as the learning rate scheduling strategy, for a total of 24 epochs.

\subsubsection{Results and Analysis}
We conduct extensive experiments on the meaningful relation prediction task and the results are reported in Table~\ref{tab:sgg_results}. The popular and classic models, such as IMP, Motifs, and RTN can produce effective values of Recall and Mean Recall. This demonstrates the effectiveness of our dataset and framework while also showing the significant challenges that exist in the proposed dataset.
Besides, VCTree develops a dynamic tree-like reasonable structure to encode visual objects and relations. It reaches a pleasing mR@50/mR@100 of 56.9\%/68.8\% on the PredCLS sub-task due to full perception of the visual context of the whole image. 
PENET solves the relation prediction problem by the prototype-representation matching approach and it reaches the impressive mR@20/mR@50 of 45.7\%/60.9\% on the PredCLS sub-task. 
SQUAT proposes a selective quad-attention network to filter the irrelevant object pairs and purify numerous candidate relations. Thus, SQUAT yields a decent R@20/R@50 of 31.5\%/39.6\% on the SGDET sub-task because the global attention mechanism enables sufficient interaction between edges and instance objects. 
In addition, there may be a lot of room for improvement because existing methods are not well adapted to model remote sensing scenarios. 
The proposed EGCA-Net introduces a RelPN module based on the self-attention mechanism to pre-select the object pairs to be predicted. This effectively eliminates some object pairs that are likely to produce confusing predictions. By extracting features with richer contextual information for pair-wise analysis and incorporating additional geometric information, EGCA-Net provides more discriminative features for the subsequent transformer-based relation predictor. As a result, EGCA-Net achieves remarkable performance across multiple subtasks, particularly excelling in the most challenging SGDET task, where it achieves the highest mR@20/mR@100/mR@500 scores among ten models, reaching 14.6\%/25.7\%/35.0\%.

\section{CONCLUSION}
 The ReCon1M dataset represents a significant advancement in the field of remote sensing imagery by providing a robust framework for developing and benchmarking scene graph generation algorithms. This dataset not only addresses the scarcity of specialized data for this domain but also enhances our understanding of complex visual scenes through detailed annotations of objects and their interrelationships. With its comprehensive coverage of diverse object and relation categories, the ReCon1M dataset is poised to facilitate a deeper understanding of remote sensing imagery and promote further innovation in scene graph generation technology. This will likely catalyze progress in various applications, from urban planning and environmental monitoring to defense and agricultural analysis, where accurate and detailed image analysis is crucial.
				
\vskip .5\baselineskip plus -1fil

\bibliographystyle{IEEEtran}
\bibliography{IEEEabrv, Example}

\end{sloppypar}
\end{document}